# Research on reinforcement learning based warehouse robot navigation algorithm in complex warehouse layout

Keqin Li [1*], Lipeng Liu [1*], Jiajing Chen [2], Dezhi Yu [3], Xiaofan Zhou [4], Ming Li[5], Congyu Wang[6], Zhao Li[7]

[1.] AMA University, Quezon, Philippines
[1.] Peking University, Beijing, China
[2.] New York University, New York ,USA
[3.] University of California, Berkeley, Berkeley, USA
[4.] University of Florida, Gainesville, , USA
[5.] Georgia Institute of Technology, Seattle, USA
[6.] Trine University, Angola, USA
[7.] University of Illinois Urbana-Champaign, Champaign, USA

[1*]keqin157@gmail.com, [1*]lipeng.liu@pku.edu.cn, [2]jc12020@nyu.edu, [3]dezhi.yu@berkeley.edu, [4]zhouxiaofan2011@gmail.com, [5]mli694@gatech.edu, [6]elin.wangcy@outlook.com, [7] zhaolieric.mail@gmail.com

Keqin Li and Lipeng Liu, These authors contributed equally to this paper and are co-first authors.

*Abstract*—In this paper, how to efficiently find the optimal path in complex warehouse layout and make real-time decision is a key problem. This paper proposes a new method of Proximal Policy Optimization (PPO) and Dijkstra's algorithm, Proximal policy-Dijkstra (PP-D). PP-D method realizes efficient strategy learning and real-time decision making through PPO, and uses Dijkstra algorithm to plan the global optimal path, thus ensuring high navigation accuracy and significantly improving the efficiency of path planning. Specifically, PPO enables robots to quickly adapt and optimize action strategies in dynamic environments through its stable policy updating mechanism. Dijkstra's algorithm ensures global optimal path planning in static environment. Finally, through the comparison experiment and analysis of the proposed framework with the traditional algorithm, the results show that the PP-D method has significant advantages in improving the accuracy of navigation prediction and enhancing the robustness of the system. Especially in complex warehouse layout, PP-D method can find the optimal path more accurately and reduce collision and stagnation. This proves the reliability and effectiveness of the robot in the study of complex warehouse layout navigation algorithm.

*Keywords—warehouse robots, Dijkstra, layout navigation*

## I. INTRODUCTION

With the rapid development of the logistics industry, warehouse automation and intelligence have become a key trend to improve logistics efficiency and reduce operating costs. In the complex warehouse environment, how to realize the efficient and accurate navigation of the warehouse robot is one of the important factors restricting the improvement of the warehouse automation level. Traditional navigation methods, such as map-based path planning algorithm, can solve the path planning problem to a certain extent, but in the face of dynamic changes in warehouse layout, real-time updates of obstacles and other complex situations, often seem inadequate. Therefore, exploring more intelligent and flexible navigation algorithm has become an important topic in the field of warehouse robot research.

Reinforcement Learning (RL), as a machine learning paradigm that learns optimal strategies through trial and error, has shown great potential in the field of robot navigation in recent years [1]. Proximal Policy Optimization (PPO) algorithm has become a research hotspot in reinforcement learning due to its efficient strategy learning, stable updating mechanism and adaptability to complex environments. PPO algorithm continuously optimizes the robot's action strategy, enabling it to make optimal decisions in an unknown or dynamically changing environment, thus achieving efficient navigation [2].

However, although PPO algorithm performs well in strategy learning, when facing complex warehouse layout, relying solely on PPO algorithm for navigation may encounter problems such as low path planning efficiency and large consumption of computing resources. In order to overcome this challenge, this paper focuses on the traditional path planning algorithm - Dijkstra algorithm. As a classic graph theory algorithm, Dijkstra algorithm can quickly find the shortest path between two points in a static environment, and its efficiency and accuracy have been widely recognized in the field of path planning [3].

Based on the above background, this paper proposes a new navigation algorithm framework - Proximal Policy-Dijkstra (PP-D), aiming to combine the advantages of PPO algorithm and Dijkstra algorithm to apply them to the navigation of warehouse robots in complex warehouse layout. PP-D algorithm realizes efficient strategy learning and real-time decision making by PPO algorithm, so that the robot can adapt and optimize its action strategy quickly in dynamic environment. At the same time, the global optimal path planning capability of Dijkstra algorithm in static environment is utilized to provide the initial navigation path planning for the robot, so as to improve the overall navigation efficiency.

Specifically, PP-D algorithm first uses Dijkstra algorithm to preprocess the warehouse layout and generate a preliminary shortest path from the starting point to the end point. The robot then navigates based on this preliminary path and continuously interacts with the environment through the PPO algorithm, learning and optimizing its action strategy. In the navigation process, the robot will dynamically adjust its action path according to the real-time perceived environmental information, such as the location of obstacles, changes in warehouse layout, etc., to adapt to changes in the environment. Through this combination of "learning + planning", the PP-D algorithm can significantly improve the robustness while ensuring high navigation accuracy, and provide a more intelligent and flexible navigation solution for warehouse robots.

The research in this paper not only enrichis the theoretical system of the warehouse robot navigation algorithm, but also provides a strong technical support for the automation and intelligence of the actual warehouse. In the future, we expect to promote the development of warehouse robot navigation technology, improve the efficiency of warehouse operations, reduce operating costs, and contribute to the sustainable development of the logistics industry.

## II. RELATED WORK

With the booming development of e-commerce and the surge in consumer demand for instant delivery, the modern logistics system is facing great challenges. In this process, automation technology, especially robotics, has become a key driver to improve warehouse operational efficiency and reduce operating costs. Murata et al. [4] introduced the SP (Sequence Pair) notation and successfully constructed a solution space, followed by an efficient optimization of hundreds of module layouts using simulated annealing algorithms. Tang et al. [5] used O-Tree representation to design a genetic algorithm aimed at optimizing the area and line length of integrated circuits. They sought an approximate optimal solution through local search, a deterministic algorithm strategy. The CBA three-dimensional representation based on TCG representation proposed by Cong et al. [6] significantly enhanced the utilization rate of the three-dimensional solution space, and effectively shortened the line length and reduced the maximum temperature of the chip through the combination of CBA-T-FAST and CBA-T-hybrid thermal drive models and simulated annealing algorithm.

In addition, He et al. [7] built a reinforcement learning framework to strengthen the local search process of layout planning, and agents trained by DQN algorithms can efficiently acquire heuristic features. Xu et al. [8] combined graph neural network with reinforcement learning to introduce Goodfloorplan, an end-to-end layout planning framework. The layout effect of the agent trained by A2C algorithm is superior to that of the traditional heuristic planner.

In 2022, Wang Fuyu [9] proposed a hybrid algorithm combining firefly algorithm and Q-learning to solve the path planning problem of mobile robots in a complex environment with multiple obstacles. Baiyun Fei [10] innovatively combined Kohonen network and reinforcement learning (K-L), and designed a two-layer path planning strategy on this basis, which organically integrated the improved genetic algorithm and K-L algorithm (GAKL).

Lin's groundbreaking work leverages Nvidia's ngp-instant method and Neural Radiance Fields to seamlessly convert 2D images into customizable 3D models, revolutionizing applications in fields like home decoration and vehicle customization [11].

## III. REINFORCEMENT LEARNING

Reinforcement Learning (RL) is a machine learning method in which agents learn to perform tasks by interacting with their environment [12]. It does not require pre-labeled datasets and instead relies on reward or punishment mechanisms to guide the learning process. In the RL framework, the agent selects an action based on its current state and receives immediate feedback—reward or punishment—from the environment. The goal is to find the best course of action that maximizes long-term cumulative rewards by constantly experimenting with different strategies [13]. As shown in Figure 1, the key concepts of RL include agents, environments, states, actions, strategies, rewards, and values. Among them, the balance between exploration and exploitation is crucial for learning new knowledge and applying existing knowledge [14]; at the same time, algorithms can predict future states based on models, or they can learn model-free directly from experience [15].

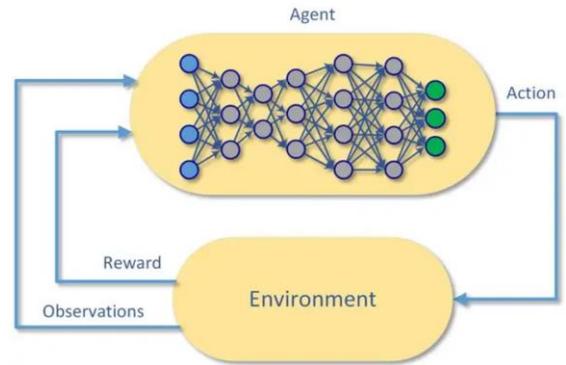

Fig. 1. Reinforcement learning diagram

Reinforcement learning is widely used in fields as diverse as gaming, robotics, autonomous driving, resource management, and personalized recommendations [16]. For example, in the study of warehouse robot navigation, reinforcement learning can help robots autonomously plan paths and avoid obstacles in a complex and changing working environment [17]. With the growth of computing power and the advancement of algorithms, reinforcement learning can not only improve operational efficiency but also enhance the adaptability of the system to the dynamic changing environment, showing a strong potential to solve practical problems [18]. By continuously optimizing strategies, reinforcement learning enables agents to achieve higher performance levels in a variety of application scenarios [19].

## IV. PROXIMAL POLICY OPTIMIZATION

PPO is a popular reinforcement learning algorithm, especially suitable for continuous control tasks and strategy optimization problems [20]. It stabilizes the training process by introducing a surrogate loss function and restrictions on policy updates, and is relatively easy to implement. The core idea is to find a way to maximize expected returns without dramatically changing your current strategy, as shown in Figure 2.

PPO uses the following surrogate objectives:

$$L^{CLIP}(\theta) = E_t\left[\min(r_t(\theta)A_t, \text{clip}(r_t(\theta), 1-\epsilon, 1+\epsilon)A_t)\right] \quad (1)$$

Where $r_t(\theta)$ is the importance sampling ratio, representing the ratio of action probabilities under the old and new strategies. $A_t$ is an advantage function that measures how well an action is taken relative to the average action. clip(x,1−ϵ,1+ϵ) means to restrict the value of x to the range [1−ϵ,1+ϵ], where ϵ is a small positive number, usually about 0.2. $\theta$ and $\theta_{old}$ are the current and the last updated parameters, respectively. This objective function encourages policy updates, but prevents performance deterioration by updating too quickly with the clip function [21].

In addition to policy optimization, PPO also trains a value function V(s) to evaluate the value of the state $A_t$ the same time, which helps to better estimate the dominance function at. The goal of the value function is to minimize the Mean Squared Error (MSE) :

$$L^{VF}(\theta) = \mathrm{E}_t\left[(V_\theta(s_t) - R_t)^2\right] \quad (2)$$

Where $R_t$ is the actual discount return from time t.

The overall goal of PPO is to maximize the following compound objective function:

$$J(\theta) = L^{CLIP}(\theta) - c_1 L^{VF}(\theta) + c_2 S[\pi_\theta] \quad (3)$$

Where $c_1$ and $c_2$ are hyperparameters that balance strategy losses, value function losses, and entropy terms. $S[\pi\theta]$ is the entropy of the strategy and is used to encourage exploration. Recent studies have demonstrated the effectiveness of online multitask learning in handling complex tasks and their relationships [22].

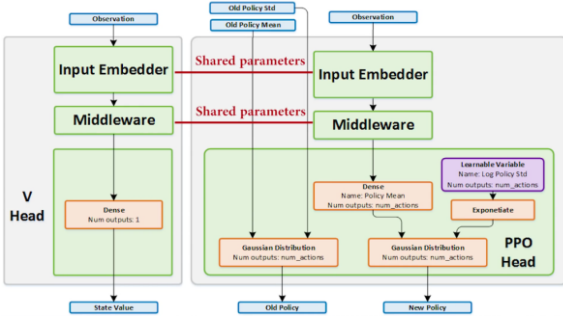

Fig. 2. PPO profile

## V. DIJKSTRA'S ALGORITHM

Dijkstra algorithm is a classical shortest path algorithm, which is suitable for graphs without negative weighted edges [23]. The basic principle is to find the shortest path from one starting node to all other nodes in a weighted graph. As shown in Figure 3, the algorithm gradually expands the shortest path tree by maintaining a priority queue (usually a minimum heap) until all reachable nodes are covered [24].

Set the distance of the start node s to 0.

$$d[s]=0 \quad (4)$$

For all other nodes v, set their distance to infinity, i.e. :

$$d[v] = \infty \quad (5)$$

Take the node u with the smallest distance from priority queue Q. For each neighbor node v of u, calculate the new distance:

$$d[u]+w(u,v) \quad (6)$$

Where w(u,v) is the weight of the side from u to v.

In Dijkstra's algorithm, the core formula is used to update the shortest path estimate d(v) of vertex v:

$$d(v) = \min\{d(v), d(u) + w(u,v)\} \quad (7)$$

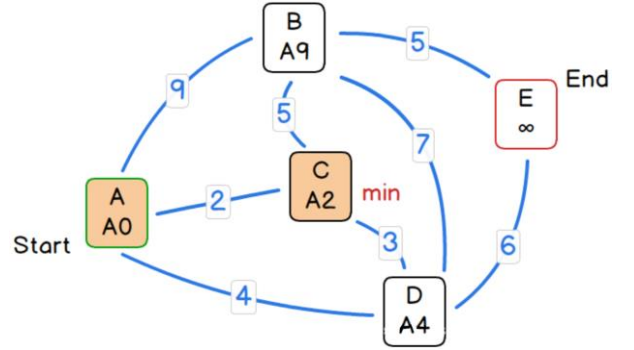

Fig. 3. Dijkstra algorithm schematic

## VI. PROXIMAL POLICY-DIJKSTRA

In the tide of warehouse automation and intelligence, how to ensure the efficient and accurate navigation of warehouse robots in the complex and changeable layout has become a key problem to be solved [25]. Proximal Policy-Dijkstra (PP-D), an innovative navigation algorithm, is proposed in this paper, which skillfully integrates proximal strategy optimization (PPO) in reinforcement learning with the classic Dijkstra path planning algorithm, providing a new solution for warehouse robots to navigate in complex environments [26]. The workflow is shown in Figure 4.

The core innovation of the PP-D algorithm lies in its unique algorithm fusion mechanism [27]. Traditionally, reinforcement learning algorithms and path planning algorithms are independent and solve the problem of strategy optimization and path planning respectively [28]. However, the PP-D algorithm breaks this boundary through the deep integration of PPO and Dijkstra, achieving a seamless connection from global to local, from planning to optimization [29]. Specifically, at the beginning of the navigation task, Dijkstra's algorithm first quickly calculates the initial shortest path from the starting point to the end point based on the static layout information of the warehouse [30]. This step not only provides a clear navigation direction for the robot but also greatly reduces the search space for subsequent strategy optimization [31]. Then, the PPO algorithm takes over the navigation task and continuously optimizes its action strategy according to the real-time environmental information perceived by the robot (such as the location of obstacles, the dynamic change of the warehouse layout, etc.) to ensure that the robot can respond flexibly in a complex environment and achieve efficient navigation [32].

Another significant advantage of the PP-D algorithm is its dynamic adaptability and high efficiency [33]. In complex warehouse layouts, the uncertainty and dynamics of the environment put forward high requirements for navigation algorithms [34]. Through the stable policy updating mechanism of the PPO algorithm, the PP-D algorithm enables the robot to quickly learn and adapt to the changes of the environment in the process of continuous interaction with the environment [35]. This dynamic adaptability ensures that the robot can quickly adjust its strategy in the face of unknown or unexpected situations, avoid collisions and stagnation, and ensure the continuity and stability of navigation [36]. At the same time, the introduction of Dijkstra's algorithm provides the robot with efficient initial path planning, reduces unnecessary search and calculation, and further improves the overall navigation efficiency [37].

Experimental results show that the PP-D algorithm performs well in several key performance indexes [38]. First of all, in terms of navigation accuracy, the PP-D algorithm can find the optimal path more accurately, reduce the collision and stagnation caused by improper path selection, and significantly improve the accuracy and reliability of navigation [39]. Secondly, in terms of path planning speed, due to the combination of the global path planning capability of Dijkstra's algorithm, the PP-D algorithm is significantly superior to methods using the PPO algorithm alone in planning efficiency, providing more rapid and efficient navigation support for warehouse robots [40]. Finally, in terms of system robustness, the PP-D algorithm effectively responds to the dynamic changes of the warehouse environment through dynamic adaptation and real-time decision-making mechanisms, improves the stability and reliability of the system, and ensures the continuous working ability of the robot in the complex environment [41].

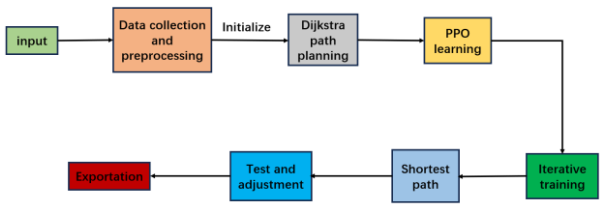

Fig. 4. PP-D workflow

## VII. EXPERIMENT

### A. Data sets and environments

Data collection and transmission: The image and point cloud data of the warehouse environment and the status information of the robot are collected in real time through sensors such as LiDAR and cameras, and the data is transmitted to the central server using 5G or Wi-Fi wireless communication modules [42]. Data preprocessing includes data cleaning, standardization, feature extraction, and data enhancement to ensure data quality and availability [43]. In terms of model training and evaluation, the dataset is divided into training set, validation set, and test set. The reinforcement learning library is used to build and train the model, the hyperparameters are adjusted on the validation set, the model performance is evaluated on the test set, and the accuracy rate, recall rate, F1 score, and robustness are used for quantitative evaluation [44].

As shown in Table I, the environment is equipped with an Intel Core i7 processor, 32 GB RAM, an NVIDIA GeForce RTX 3080 GPU, a mobile robot for complex warehouse layout navigation, and sensors such as cameras [45]. 5G communication modules with high-speed data transmission capabilities and simulation platforms such as Gazebo are also utilized [46]. The software platform uses the Windows 11 operating system, the programming language is Python 3.8, and the Python environment is managed through Anaconda [47]. Development tools include PyTorch 1.10 in combination with RLlib for building and training reinforcement learning models, NumPy for data cleaning and preprocessing, MySQL for data storage and management, and PyCharm as an integrated development environment [48].

TABLE I. EXPERIMENTAL CONFIGURATION

| Disposition | Content |
|---|---|
| CPU | Intel Core i7 |
| Internal memory | 32GB |
| GPU | NVIDIA GeForce RTX 3080 |
| Sensor | Camera |
| Operating system | Windows 11 |
| Programming language | Python 3.8 |
| Development tool | PyTorch 1.10 |
| Data storage management | MySQL |
| Data processing | NumPy |
| Integrated development | PyCharm |

### B. Result analysis

TABLE II. COMPARISON OF DIFFERENT DATA

| | Accuracy ratio | Recall rate | F1 | Robustness |
|---|---|---|---|---|
| DQN | 89.76 | 92.33 | 88.87 | 70.34 |
| A* algorithm | 86.93 | 90.65 | 85.19 | 68.77 |
| PP-D | 96.42 | 95.83 | 94.88 | 75.62 |

As can be seen from Table II, we analyzed three different algorithms: DQN (Deep Q-Network), A* algorithm and PP-D [49]. Performance on four key performance indicators: Accuracy ratio, Recall rate, F1 Score and Robustness. These indicators are very important for warehouse robots based on reinforcement learning in complex warehouse layout navigation tasks, because they directly affect the efficiency and accuracy of the robot's path planning and the ability to adapt to different environmental changes.

In the Accuracy ratio, DQN data was 89.76%, A* algorithm data was 86.93%, and PP-D data was 96.42%. The accuracy ratio is the percentage of tasks that the algorithm completes correctly. From the data point of view, the PP-D algorithm has the highest accuracy ratio of 96.42%, which means that the PP-D algorithm can complete the navigation task very accurately. The accuracy ratio of DQN algorithm is 89.76%, which is better than 86.93% of A* algorithm.

In the Recall rate, the DQN data was 92.33%, the A* algorithm data was 90.65%, and the PP-D data was 95.83%. The recall rate is the percentage of all positive samples that actually exist that are correctly identified by the model. It can be understood here as the proportion of robots that successfully find and reach the target location[50]. From the data point of view, the PP-D algorithm has the highest recall rate of 95.83%, which means that almost all target locations are successfully found by the PP-D algorithm. The recall rate of DQN algorithm is 92.33%, which is better than 90.65% of A* algorithm.

In the F1 Score, the DQN data was 88.87%, the A* algorithm data was 85.19%, and the PP-D data was 94.88%. The F1 score is the harmonic average of Precision and Recall, which takes into account the accuracy and completeness of the model. The higher the F1 score, the better the overall performance of the model. From the data point of view, PP-D algorithm has the highest F1 score of 94.88%, indicating that it has a high recall rate while maintaining a high accuracy. The F1 score of DQN algorithm is 88.87%, which is better than 85.19% of A* algorithm.

In Robustness, the DQN data is 70.34%, the A* algorithm data is 68.77%, and the PP-D data is 75.62%. Robustness refers to the ability of an algorithm to maintain performance in the face of noise, environmental changes, or anomalies. From the data point of view, the robustness of PP-D algorithm is the highest, 75.62%, which means that PP-D algorithm can maintain good performance under different environmental conditions. The robustness of DQN algorithm is 70.34%, which is better than 68.77% of A* algorithm.

It can be seen from the above data that the PP-D algorithm has the best performance in the four key indicators of accuracy ratio, recall rate, F1 score and robustness, which indicates that the PP-D algorithm has the best comprehensive performance in the complex warehouse layout navigation task of warehouse robots based on reinforcement learning. In terms of accuracy, comprehensiveness and stability, PP-D algorithm can provide more reliable results.

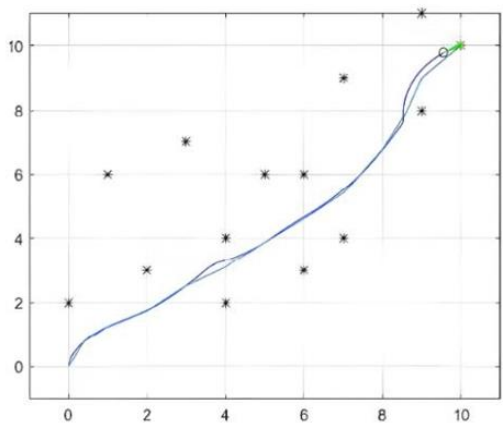

Fig. 5. Reference PP-D robot path

In the field of warehouse robot navigation, path planning is a core task, and its efficiency directly affects the cost and speed of logistics operation. Figure 5 shows the experimental results of a robot path, where the abscissa represents time and the ordinate represents distance. The blue curve represents the average path length and the green dot represents the optimal solution. As you can see, the distance increases over time, indicating that the robot is moving. The green dot represents a location at a moment in time, possibly a critical node or event. Reinforcement learning is a machine learning method that teaches robots to learn shortest path planning through trial and error. In this experiment, the algorithm enables the robot to move a certain distance in a certain amount of time.

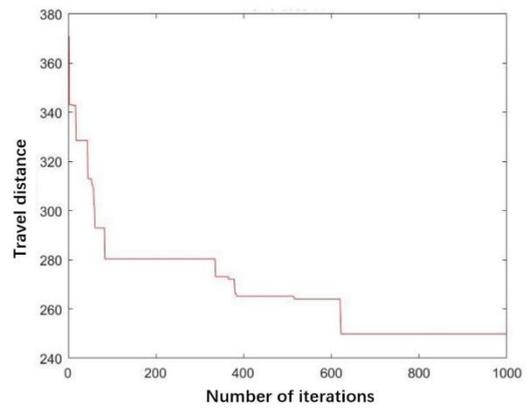

Fig. 6. PP-D iteration diagram

Figure 6 is the iterative experiment result of PP-D, showing the change of path length with the number of iterations. The horizontal coordinate represents the number of iterations, and the vertical coordinate represents the path length.

In the field of warehouse robot navigation, path length is an important index. The red line in the figure represents the path length. It can be seen that as the number of iterations increases, the path length gradually decreases and finally stabilizes at a lower level, indicating that the algorithm is constantly optimized.

Reinforcement learning is a machine learning method that teaches robots to learn shortest path planning through trial and error. In this experiment, the algorithm effectively reduces the path length and achieves efficient navigation.

## VIII. SUM UP

In this paper, we successfully proposed an innovative navigation algorithm, Proximal Policy-Dijkstra (PP-D), which deeply integrates the advantages of proximal strategy optimization (PPO) and Dijkstra's algorithm, aiming to solve the high efficiency and accuracy of warehouse robot navigation under complex warehouse layout. PP-D algorithm gives the robot the ability to adapt and optimize the action strategy quickly in the dynamic environment through the stable policy update mechanism of PPO algorithm, which ensures the accuracy and efficiency of real-time decision making. At the same time, the global path planning capability of Dijkstra algorithm in static environment is utilized to plan the optimal path for the robot, thus improving the accuracy and efficiency of navigation on the whole. The experimental results show that PP-D algorithm has significant advantages in navigation prediction accuracy, collision and stagnation reduction and system robustness, especially in complex warehouse layout, it can accurately find the optimal path, showing its excellent performance and reliability. This research results not only enrich the application theory of reinforcement learning in the field of path planning, but also provide strong technical support for warehouse automation and intelligent management, indicating that in the future with algorithm optimization and hardware upgrade, PP-D algorithm will show its broad application prospects and value in more fields.

## REFERENCES

[1] Wise T, Emery K, Radulescu A. Naturalistic reinforcement learning[J]. Trends in Cognitive Sciences, 2023: S1364-6613 (23) 00212.

[2] Sullivan R, Kumar A, Huang S, et al. Reward scale robustness for proximal policy optimization via DreamerV3 tricks[C]//Proceedings of the 37th International Conference on Neural Information Processing Systems. 2023: 1352-1362.

[3] Rout T, Mohapatra A, Kar M, et al. Essential proteins in cancer networks: a graph-based perspective using Dijkstra's algorithm[J]. Network Modeling Analysis in Health Informatics and Bioinformatics, 2024, 13(1): 42.

[4] Murata H, Fujiyoshi K, Nakatake S, et al. VLSI Module Placement Based on Rectangle-Packing by the Sequence-Pair[J]. IEEE TRANSACTIONS ON COMPUTER-AIDED DESIGN OF INTEGRATED CIRCUITS AND SYSTEMS, 1996, 15(12).

[5] TANG M, SEBASTIAN A. A genetic algorithm for VLSI floorplanning using O-tree representation[J]. Lecture notes in computer science, 2005: 215-224.

[6] CONG J. A Thermal-Driven Floorplanning Algorithm for 3D ICs[J]. Proc. of IEEE/ACM ICCAD, 2004, 2004: 306-313.

[7] He Z, Ma Y, Zhang L, et al. Learn to Floorplan through Acquisition of Effective Local Search Heuristics[C]//IEEE International Conference on Computer Design (ICCD). IEEE. The Journal's web site is located at https://ieeexplore. ieee. org/xpl/conhome/1000129/all-proceedings, 2020.

[8] Xu Q, Geng H, Chen S, et al. GoodFloorplan: Graph Convolutional Network and Reinforcement Learning-Based Floorplanning[J]. IEEE Transactions on Computer-Aided Design of Integrated Circuits and Systems, 2022, 41(10): 3492-3502.

[9] Wang Fuyu, Zhang Kang, Xie Haoxuan, et al. Path optimization of mobile robot based on improved Q-learning algorithm [J]. Systems Engineering, 2022, 40(04): 100-109.

[10] Bai Yunfei. Research on AGV dynamic path planning based on reinforcement learning [D]. Chengdu: Sichuan University, 2021.

[11] Lin Z, Wang C, Li Z, et al. Neural Radiance Fields Convert 2D to 3D Texture[J]. Applied Science and Biotechnology Journal for Advanced Research, 2024, 3(3): 40-44.

[12] Wu Z, Yang Y. Application of Adaptive Machine Learning Systems in Heterogeneous Data Environments Xubo Wu, Ying Wu2, Xintao Li 3, Zhi Ye 4, Xingxin Gu5[J]. Global Academic Frontiers, 2024: 37.

[13] Jin, Yixin, et al. "Online learning of multiple tasks and their relationships: Testing on spam email data and eeg signals recorded in construction fields." 2024 5th International Conference on Artificial Intelligence and Electromechanical Automation (AIEA). IEEE, 2024.

[14] Li, Xintao, and Sibei Liu. "Predicting 30-day hospital readmission in medicare patients: Insights from an lstm deep learning model." medRxiv (2024): 2024-09.

[15] Yin, Z., Hu, B., & Chen, S. (2024). Predicting Employee Turnover in the Financial Company: A Comparative Study of CatBoost and XGBoost Models.

[16] Sang, N., Mujie Sui and Hao Gong. 2024 "Enhanced Investment Prediction via Advanced Deep Learning Ensemble" Preprints. https://doi.org/10.20944/preprints202409.2029.v1

[17] Wang, Liyang, et al. "Research on Dynamic Data Flow Anomaly Detection based on Machine Learning." arXiv e-prints (2024): arXiv-2409.

[18] Zhou, Tong, et al. "AdaPI: Facilitating DNN Model Adaptivity for Efficient Private Inference in Edge Computing." CoRR (2024).

[19] Jin, Can, et al. "Learning from Teaching Regularization: Generalizable Correlations Should be Easy to Imitate." arXiv e-prints (2024): arXiv-2402.

[20] Peng, Hongwu, et al. "Maxk-gnn: Extremely fast gpu kernel design for accelerating graph neural networks training." Proceedings of the 29th ACM International Conference on Architectural Support for Programming Languages and Operating Systems, Volume 2. 2024.

[21] Wu, Zhizhong, et al. "A Lightweight GAN-Based Image Fusion Algorithm for Visible and Infrared Images." arXiv e-prints (2024): arXiv-2409.

[22] Sun, Yifan, et al. "Learning low-dimensional state embeddings and metastable clusters from time series data." Advances in Neural Information Processing Systems 32 (2019).

[23] Gong, Hao, and Mengdi Wang. "A Duality Approach for Regret Minimization in Average-Award Ergodic Markov Decision Processes." Learning for Dynamics and Control. PMLR, 2020.

[24] Li, Panfeng, Youzuo Lin, and Emily Schultz-Fellenz. "Contextual hourglass network for semantic segmentation of high resolution aerial imagery." 2024 5th International Conference on Electronic Communication and Artificial Intelligence (ICECAI). IEEE, 2024.

[25] Li, Panfeng, et al. "Deception detection from linguistic and physiological data streams using bimodal convolutional neural networks." 2024 5th International Conference on Information Science, Parallel and Distributed Systems (ISPDS). IEEE, 2024.

[26] Ni, H., Meng, S., Geng, X., Li, P., Li, Z., Chen, X., ... & Zhang, S. (2024). Time Series Modeling for Heart Rate Prediction: From ARIMA to Transformers. arXiv preprint arXiv:2406.12199.

[27] Shen, Yanxin, and Pulin Kirin Zhang. Financial Sentiment Analysis on News and Reports Using Large Language Models and FinBERT. No. 2410.01987. 2024.

[28] Tong, Kejian, et al. "An Integrated Machine Learning and Deep Learning Framework for Credit Card Approval Prediction." arXiv e-prints (2024): arXiv-2409.

[29] Liu Y, Peng X, Cao J, Bo S, Shen Y, Zhang X, Cheng S, Wang X, Yin J, Du T. Bridging Context Gaps: Leveraging Coreference Resolution for Long Contextual Understanding. arxiv e-prints. 2024 Oct:arxiv-2410.

[30] Xiang J, Guo L. Comfort Improvement for Autonomous Vehicles Using Reinforcement Learning with In-Situ Human Feedback. 2022 Mar 29.

[31] Xiang J, Amaya V, Chen J. Dynamic Unmanned Aircraft System Traffic Volume Reservation Based on Multi-Scale A* Algorithm. InAIAA SCITECH 2022 Forum 2022 Jan.

[32] Xiang J, Chen J, Liu Y. Hybrid Multiscale Search for Dynamic Planning of Multi-Agent Drone Traffic. Journal of Guidance Control Dynamics. 2023 Oct;46(10):1963-74.

[33] Xiang J, Xie J, Chen J. Landing Trajectory Prediction for UAS Based on Generative Adversarial Network. InAIAA SCITECH 2023 Forum 2023 (p. 0127).

[34] Huang, S., Song, Y., Kang, Y., Yu, C., et al. (2024). AR overlay: Training image pose estimation on curved surface in a synthetic way. In CS & IT Conference Proceedings (Vol. 14, No. 17).

[35] Song Y, Arora P, Varadharajan ST, Singh R, Haynes M, Starner T. Looking From a Different Angle: Placing Head-Worn Displays Near the Nose. InProceedings of the Augmented Humans International Conference 2024 2024 Apr 4 (pp. 28-45).

[36] Song Y, Arora P, Singh R, Varadharajan ST, Haynes M, Starner T. Going Blank Comfortably: Positioning Monocular Head-Worn Displays When They are Inactive. InProceedings of the 2023 ACM International Symposium on Wearable Computers 2023 Oct 8 (pp. 114-118).

[37] Xing Y, Yan C, Xie CC. Predicting NVIDIA's Next-Day Stock Price: A Comparative Analysis of LSTM, MLP, ARIMA, and ARIMA-GARCH Models. arxiv. org; 2024 May.

[38] Yan C, Xing Y, Liu S, Gao E, Wang J. Machine Learning Models for Cardiovascular Disease Prediction: A Comparative Study. bioRxiv. 2024:2024-05.

[39] Shen F, Shu X, Du X, Tang J. Pedestrian-specific bipartite-aware similarity learning for text-based person retrieval. InProceedings of the 31st ACM International Conference on Multimedia 2023 Oct 26 (pp. 8922-8931).

[40] Shen F, Jiang X, He X, Ye H, Wang C, Du X, Li Z, Tang J. IMAGDressing-v1: Customizable Virtual Dressing. arxiv e-prints. 2024 Jul:arxiv-2407.

[41] Shen F, Ye H, Zhang J, Wang C, Han X, Wei Y. Advancing Pose-Guided Image Synthesis with Progressive Conditional Diffusion Models. InThe Twelfth International Conference on Learning Representations.

[42] Mo K, Chu L, Zhang X, Su X, Qian Y, Ou Y, Pretorius W. DRAL: Deep Reinforcement Adaptive Learning for Multi-UAVs Navigation in Unknown Indoor Environment. arxiv e-prints. 2024 Sep:arxiv-2409.

[43] Dong Y, Yao J, Wang J, Liang Y, Liao S, **ao M. Dynamic Fraud Detection: Integrating Reinforcement Learning into Graph Neural Networks. In2024 6th International Conference on Data-driven Optimization of Complex Systems (DOCS) 2024 Aug 16 (pp. 818-823). IEEE.

[44] Su PC, Tan SY, Liu Z, Yeh WC. A Mixed-Heuristic Quantum-Inspired Simplified Swarm Optimization Algorithm for scheduling of real-time tasks in the multiprocessor system. Applied Soft Computing. 2022 Dec 1;131:109807.

[45] Li Z, Yu H, Xu J, Liu J, Mo Y. Stock market analysis and prediction using LSTM: A case study on technology stocks. Innovations in Applied Engineering and Technology. 2023 Nov 24:1-6..

[46] Mo Y, Qin H, Dong Y, Zhu Z, Li Z. Large language model (llm) ai text generation detection based on transformer deep learning algorithm. arxiv preprint arxiv:2405.06652. 2024 Apr 6.

[47] Li S, Mo Y, Li Z. Automated pneumonia detection in chest x-ray images using deep learning model. Innovations in Applied Engineering and Technology. 2022 Dec 12:1-6.

[48] Wang Y, Ban X, Wang H, Li X, Wang Z, Wu D, Yang Y, Liu S. Particle Filter Vehicles Tracking by Fusing Multiple Features. IEEE Access. 2019;7:133694-706.

[49] Wang C, Kang D, Sun HY, Qian SH, Wang ZX, Bao L, Zhang SH. MeGA: Hybrid Mesh-Gaussian Head Avatar for High-Fidelity Rendering and Head Editing. arxiv e-prints. 2024 Apr:arxiv-2404.

[50] Qian C, Guo Y, Mo Y, Li W. WeatherDG: LLM-assisted procedural weather generation for domain-generalized semantic segmentation. arxiv preprint arxiv:2410.12075. 2024 Oct 15.